\documentclass[10pt,twocolumn,letterpaper]{article}

\usepackage[margin=0.75in]{geometry}
\usepackage{times}
\usepackage{graphicx}
\usepackage{amsmath,amssymb}
\usepackage{booktabs}
\usepackage{multirow}
\usepackage{array}
\usepackage{enumitem}
\usepackage{caption}
\usepackage{placeins}
\usepackage{float}
\usepackage{xcolor}
\usepackage{tikz}
\usetikzlibrary{arrows.meta,positioning,fit,shapes.geometric,calc}
\usepackage[hidelinks]{hyperref}

\setlist[itemize]{leftmargin=1.2em,itemsep=1pt,topsep=2pt}

\title{\vspace{-0.45in}\bfseries Turbulence-Robust Dynamic Object Segmentation with Multi-Signal Priors and SAM2 Refinement\\
{\large Technical Report for CVPR 2026 UG2+ Challenge Track 3: DOST}}

\author{Bolian Peng$^{1}$ \qquad Ying Tang$^{1}$ \qquad Xu Liu$^{1}$ \qquad Long Sun$^{1}$ \qquad Xiaoqiang Lu$^{1}$\\
$^{1}$Xidian University\\
{\tt\small janeyo0910@email.com}
}

\date{}

\begin{document}
\maketitle
\thispagestyle{empty}

\begin{abstract}
This technical report presents our solution for the CVPR 2026 UG2+ Challenge Track 3: Dynamic Object Segmentation in Turbulence (DOST). We design a training-free multi-signal segmentation pipeline that combines pretrained motion estimation, self-supervised semantic priors, background anomaly modeling, manually calibrated proposal fusion, and SAM2-based mask refinement. The method uses RAFT for dense motion responses, DINOv2 for semantic objectness priors, ViBe for training-free background modeling, and pretrained SAM2 for box-prompt mask refinement.

Instead of optimizing an end-to-end segmentation network, our system operates entirely in inference mode. This design is suitable for the DOST setting, where severe atmospheric turbulence produces pseudo-motion, blur, and intermittent target visibility, making a single motion cue unreliable. The final submitted masks are evaluated by the official leaderboard, which reports 0.425041 mIoU and 0.457206 mDice. Since no task-specific model training or fine-tuning is performed, stronger learned temporal association, adaptive proposal selection, or task-specific adaptation may further improve the system.
\end{abstract}

\section{Introduction}
The Dynamic Object Segmentation in Turbulence (DOST) challenge focuses on per-frame segmentation of dynamic foreground objects in real-world long-range videos. The track uses the DOST dataset introduced in~\cite{dost}, where each sequence contains turbulence-degraded frames and pixel-level masks for moving objects. Atmospheric turbulence introduces spatially varying geometric distortion, non-uniform blur, and temporally unstable appearance changes. These effects distinguish DOST from conventional video segmentation: apparent background displacement caused by turbulence can be difficult to separate from actual object motion.

This setting makes motion-based segmentation particularly challenging. A turbulent background may exhibit non-rigid pseudo-motion even when the scene is physically static, while small foreground targets can become blurred, intermittent, or visually mixed with local background fluctuations. Consequently, methods that rely on frame difference or optical-flow magnitude alone tend to produce fragmented foreground responses or activate background regions.

To handle these difficulties, we adopt a fully training-free inference pipeline that uses several complementary evidence sources. RAFT provides dense optical-flow responses, skip-frame flow increases the visibility of weak motion, DINOv2 supplies a semantic objectness prior, ViBe models background anomalies, and SAM2 refines coarse proposals through box prompts. All deep models are used with pretrained weights, and no DOST-specific network training or fine-tuning is performed. The design is not intended to be a purely end-to-end architecture. Instead, it prioritizes stability under severe turbulence and allows different cues to compensate for each other's failure cases.

The contributions of this report are summarized as follows:
\begin{itemize}
    \item We present a training-free dynamic object segmentation pipeline that combines motion, semantic, background-anomaly, temporal, and prompt-based cues for turbulence-degraded videos.
    \item We introduce a DINO-guided proposal stage that suppresses part of the turbulence-induced pseudo-motion while preserving semantically coherent object regions.
    \item We incorporate skip-frame optical flow and ViBe background modeling to improve weak-motion detection and anomaly discrimination in difficult scenes.
    \item We use SAM2 box-prompt refinement with isolated-box filtering and temporal proposal recovery to convert noisy coarse proposals into more coherent object-level masks.
\end{itemize}

\section{Related Work}
\subsection{Motion Estimation and Motion Segmentation}
Motion cues have long been used for dynamic object segmentation, since independently moving objects usually exhibit displacement patterns that differ from the background. Classical approaches rely on frame differencing, optical flow, trajectory clustering, or background compensation to separate foreground regions from the scene. Recent learning-based optical-flow methods, represented by RAFT~\cite{raft}, further improve dense correspondence estimation by iteratively updating flow fields over a correlation volume.

Atmospheric turbulence, however, weakens a key assumption behind many motion segmentation methods: the background is no longer static or globally smooth. In long-range turbulence-degraded videos, local background structures may undergo non-rigid and spatially varying fluctuations, producing strong pseudo-motion across the image. Therefore, an optical-flow magnitude map can provide useful motion evidence, but it is often too noisy to serve as a foreground mask by itself. This motivates the use of optical flow as one cue within a broader proposal-generation pipeline.

\subsection{Semantic Priors from Self-Supervised Visual Features}
Self-supervised visual representation learning has shown strong transferability to downstream recognition and localization tasks. Models such as DINOv2~\cite{dinov2} learn object-level feature consistency from large-scale unlabeled images, making their representations less sensitive to local pixel noise than raw appearance or short-term displacement. This property is useful for weak localization in degraded videos, where apparent motion can be unreliable.

Semantic features alone are not sufficient for dynamic object segmentation because they do not explicitly indicate whether an object is moving. Nevertheless, they can help suppress background regions that contain strong turbulence-induced motion but lack coherent object-level structure. For this reason, semantic priors are naturally complementary to motion proposals in DOST scenes.

\subsection{Background Modeling and Temporal Filtering}
Background subtraction is another common route for foreground detection. Instead of estimating object motion directly, these methods model the normal appearance distribution of each pixel or region over time and mark deviations as foreground. ViBe~\cite{vibe} is a representative sample-based background modeling method that maintains historical samples for each pixel and identifies foreground regions according to appearance inconsistency. Its training-free design makes it attractive when supervised data are limited.

For turbulence-degraded videos, background modeling complements optical flow. Real moving targets often produce persistent temporal deviations, whereas turbulence artifacts tend to be fragmented and unstable. At the same time, background subtraction can fail when targets are extremely small, slow-moving, or visually similar to the background. It is therefore more effective when combined with motion and semantic cues rather than used as an independent solution.

\subsection{Prompt-Based Mask Refinement}
Prompt-based segmentation models, including SAM~\cite{sam} and SAM2~\cite{sam2}, provide a practical way to refine coarse object proposals when reliable prompts are available. Given point, box, or mask prompts, these models can generate object-level masks with better boundary quality than many task-specific coarse proposal methods. SAM2 further extends this paradigm to video data and improves the handling of temporal mask propagation.

The main difficulty in DOST is not only mask generation, but also prompt generation. Turbulence-induced pseudo-motion can lead to inaccurate point prompts or oversized noisy regions, causing the segmentation model to activate background structures. Compared with point prompts or direct video propagation, box prompts are more stable under turbulence because they spatially constrain the refinement area and reduce the risk of activating background pseudo-motion. Our solution therefore first constructs coarse proposals by fusing motion, semantic, and background signals, and then uses the resulting connected components as box prompts for SAM2 refinement.

\section{Method}
\subsection{Overall Framework}
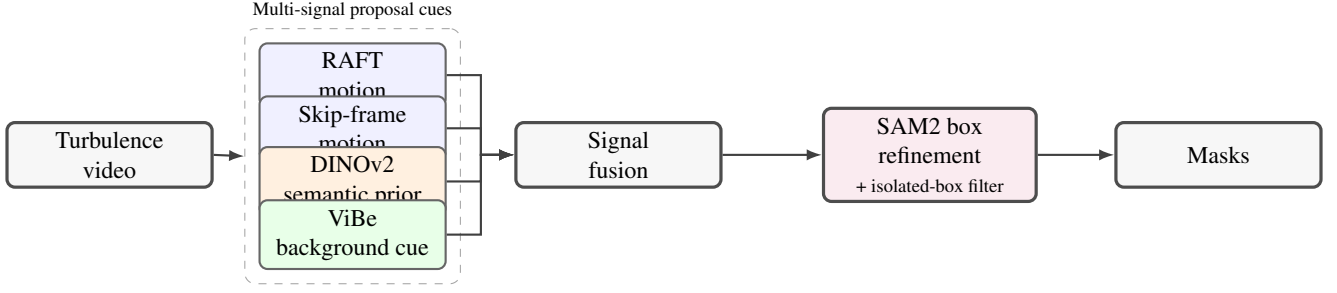
\begin{figure*}[t]
\centering
\resizebox{0.98\textwidth}{!}{%
\begin{tikzpicture}[
    font=\small,
    block/.style={rectangle, rounded corners=3pt, draw=black!70, very thick, align=center, minimum height=0.85cm, minimum width=2.70cm, fill=gray!6},
    cue/.style={rectangle, rounded corners=3pt, draw=black!60, thick, align=center, minimum height=0.66cm, minimum width=2.45cm, fill=blue!6},
    sem/.style={rectangle, rounded corners=3pt, draw=black!60, thick, align=center, minimum height=0.66cm, minimum width=2.45cm, fill=orange!12},
    bg/.style={rectangle, rounded corners=3pt, draw=black!60, thick, align=center, minimum height=0.66cm, minimum width=2.45cm, fill=green!9},
    refine/.style={rectangle, rounded corners=3pt, draw=black!70, very thick, align=center, minimum height=0.85cm, minimum width=2.80cm, fill=purple!8},
    group/.style={rectangle, rounded corners=4pt, draw=black!35, dashed, inner sep=0.18cm},
    arrow/.style={-{Latex[length=2.0mm]}, thick, draw=black!75}
]
\node[block] (input) at (0,0) {Turbulence\\video};

\node[cue] (raft) at (3.2,1.05) {RAFT\\motion};
\node[cue] (skip) at (3.2,0.35) {Skip-frame\\motion};
\node[sem] (dino) at (3.2,-0.35) {DINOv2\\semantic prior};
\node[bg] (vibe) at (3.2,-1.05) {ViBe\\background cue};

\node[group, fit=(raft)(skip)(dino)(vibe), label={[font=\scriptsize]above:Multi-signal proposal cues}] (cuebox) {};

\node[block] (fusion) at (6.7,0) {Signal\\fusion};

\node[refine, minimum width=2.8cm] (sam) at (10.8,0) 
{SAM2 box\\refinement\\
{\scriptsize + isolated-box filter}};

\node[block] (out) at (14.6,0) {Masks};

\draw[arrow] (input.east) -- (cuebox.west);
\draw[arrow] (raft.east) -- ++(0.45,0) |- (fusion.west);
\draw[arrow] (skip.east) -- ++(0.45,0) |- (fusion.west);
\draw[arrow] (dino.east) -- ++(0.45,0) |- (fusion.west);
\draw[arrow] (vibe.east) -- ++(0.45,0) |- (fusion.west);
\draw[arrow] (fusion) -- (sam);
\draw[arrow] (sam) -- (out);
\end{tikzpicture}%
}
\caption{Overview of the proposed training-free DOST framework. Pretrained RAFT, DINOv2, and SAM2 models are used without DOST-specific training. Multi-signal proposal cues are fused and refined by SAM2 box prompts with isolated-box filtering to produce the final dynamic object masks.}
\label{fig:pipeline}
\end{figure*}

Figure~\ref{fig:pipeline} illustrates the proposed framework. Given an input video, we first compute dense optical flow and derive a RAFT motion confidence map. In parallel, skip-frame flow is estimated to strengthen weak displacement, DINOv2 features are used to obtain semantic objectness priors, and ViBe provides a background anomaly response. These maps are normalized and fused into a coarse proposal. Connected components from the proposal are then converted into box prompts for SAM2, followed by isolated-box filtering. All modules are adopted from available pretrained models, and no model is trained or fine-tuned on the DOST dataset.

\subsection{RAFT Motion Proposal}
We use RAFT to estimate dense optical flow between adjacent frames. The flow magnitude is normalized into a motion confidence map, and connected-component filtering removes small isolated fragments. This cue is useful when the foreground target has clear displacement, but it is also sensitive to turbulence-induced background motion. We therefore regard the RAFT response as a lower-bound motion cue rather than a complete segmentation result.

\subsection{DINOv2 Semantic Prior}
Optical flow measures displacement rather than objectness. To introduce a more stable spatial prior, we extract DINOv2 features and construct a weak semantic objectness map. The prior reweights the motion confidence map instead of replacing it. Regions that are both motion-consistent and semantically coherent receive higher confidence, whereas scattered turbulent responses are suppressed. This DINO-guided proposal is used as the shared base for subsequent modules.

\subsection{Skip-Frame Motion Enhancement}
Adjacent-frame flow can be weak for slow-moving objects. We therefore compute optical flow over a larger temporal gap:
\begin{equation}
    F_{t \rightarrow t+k} = \operatorname{RAFT}(I_t, I_{t+k}),
\end{equation}
where $k$ is the skip interval. The longer interval amplifies true object displacement and improves the detection of slow targets, especially in vibration-dominated sequences where adjacent-frame responses are noisy.

\subsection{ViBe Background Anomaly Modeling}
ViBe is used as a lightweight background model. For each pixel, it maintains historical appearance samples and marks the current observation as foreground when it deviates from the background distribution. The resulting anomaly map complements optical flow: RAFT emphasizes displacement, whereas ViBe emphasizes persistent deviation from normal appearance. This is useful for small objects and stationary-to-moving targets, but it can also introduce false positives when background appearance changes abruptly.

\subsection{Multi-Signal Fusion}
The normalized RAFT map $M_t$, skip-frame map $M^{\mathrm{skip}}_t$, DINO semantic prior $P^{\mathrm{sem}}_t$, and ViBe anomaly response $B_t$ are fused into a proposal score:
\begin{equation}
    S_t = \alpha M_t + \beta M^{\mathrm{skip}}_t + \gamma P^{\mathrm{sem}}_t + \delta B_t.
\end{equation}
The fusion branch provides the coarse proposal score used for subsequent box-prompt generation. In our implementation, the scalar thresholds, fusion weights, and box-selection parameters are selected through visual inspection of intermediate proposals during the development workflow. For example, overly permissive thresholds may turn large background regions into foreground, while overly strict thresholds may remove tiny or weakly moving targets. The selected parameters are fixed before SAM2 refinement and are not adjusted frame by frame during inference. This manual proposal calibration is used to handle large scene-level variations in object scale, turbulence strength, and motion magnitude.

\subsection{SAM2 Box-Prompt Refinement and Isolated-Box Filtering}
After the fused proposal stage, the coarse mask is binarized and converted into connected components. Each valid component is expanded into a compact bounding box and passed to pretrained SAM2 as a box prompt. In this branch, SAM2 is used only as an inference-time boundary refinement module. It converts noisy proposal regions into more coherent object-level masks without any additional training or fine-tuning on DOST.

We further apply an isolated-box filter to suppress temporally inconsistent false positives. For each candidate box, we compare its spatial overlap with boxes in neighboring frames. Boxes with little overlap to both the previous and next frames are treated as short-lived turbulence artifacts and are removed before mask generation. Since strict filtering may also remove valid objects that appear briefly or become weak in long-range frames, a lightweight temporal proposal recovery step is used for difficult cases. In the bird sequence, reliable nearby detections are propagated to recover short missing appearances. In the aeroplane sequence, boxes from neighboring reliable frames are propagated to tail frames when the coarse proposal becomes incomplete.

\section{Experiments}
\subsection{Dataset and Evaluation}
The DOST benchmark~\cite{dost} evaluates dynamic object segmentation in turbulence-degraded videos. The labeled validation subset covers different failure modes, including tiny flying targets, long-range people and vehicles, camera vibration, stationary-to-moving cars, and blurred aircraft under strong atmospheric distortion. We report the official mean Intersection-over-Union (mIoU) and mean Dice (mDice) metrics released by the leaderboard.

\begin{figure*}[t]
\centering
\includegraphics[width=0.98\textwidth]{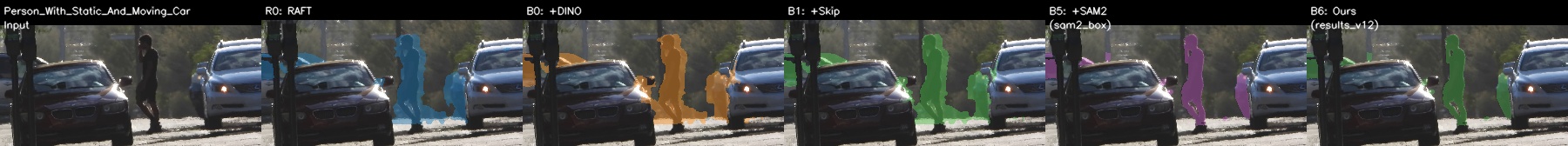}
\vspace{0.4em}
\includegraphics[width=0.98\textwidth]{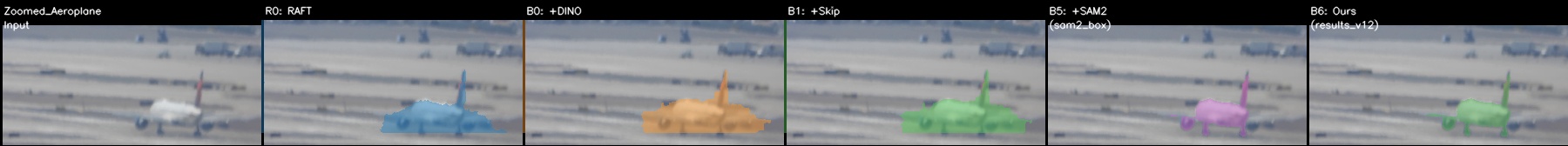}
\caption{Qualitative progression of the proposed pipeline. The first row shows \textit{Person With Static And Moving Car}, and the second row shows \textit{Zoomed Aeroplane}. The columns show the input frame, RAFT proposal, DINO-guided proposal, skip-frame proposal, SAM2 box refinement, and the final result.}
\label{fig:qual_progression}
\end{figure*}

\subsection{Inference Configuration and Workflow}
The system is implemented as a training-free modular inference pipeline. We use Python 3.12, PyTorch 2.5.1 with CUDA 12.4, and the official SAM2 implementation. All experiments are conducted on a single NVIDIA A800 80GB GPU. Since the pipeline does not involve back-propagation or model fine-tuning, its memory requirement is mainly determined by inference resolution, optical-flow computation, and SAM2 refinement. In practice, smaller GPUs can also be used by reducing the processing resolution or running the stages sequentially.

The experimental workflow follows the order of the proposed pipeline. We first compute RAFT optical flow, skip-frame flow, DINOv2 semantic priors, and ViBe anomaly responses. The resulting score maps are then inspected to select fixed thresholds, fusion weights, and box-selection parameters before the SAM2 refinement stage. This manual proposal calibration is based on visual examination of intermediate masks and typical failure cases, such as thresholds that activate most of the background or remove small foreground targets. After the parameters are fixed, the pipeline is run in inference mode to generate final masks. No DOST-specific model training or fine-tuning is performed.

\subsection{Quantitative Results}
Table~\ref{tab:official} reports the official leaderboard evaluation scores of our submitted masks. This report uses the official released metrics as the only quantitative performance numbers. The final submitted system obtains 0.425041 mIoU and 0.457206 mDice. The per-video scores show that extremely small, unstable, and long-range targets remain difficult under severe turbulence.

\begin{table}[H]
\centering
\caption{Official released evaluation results of the final submitted masks on the DOST validation subset. These scores are the only quantitative results reported in this technical report.}
\label{tab:official}
\scriptsize
\setlength{\tabcolsep}{3pt}
\begin{tabular}{p{0.58\columnwidth}cc}
\toprule
Video & mIoU $\uparrow$ & mDice $\uparrow$ \\
\midrule
Bird Behind Pillar & 0.3327 & 0.3995 \\
People Using Mobile II & 0.4807 & 0.4902 \\
Person With Static And Moving Car & 0.4456 & 0.4712 \\
Vibration Test & 0.4590 & 0.4786 \\
White Car Stop Sign & 0.4766 & 0.4880 \\
Zoomed Aeroplane & 0.3557 & 0.4157 \\
\midrule
Final evaluation & \textbf{0.425041} & \textbf{0.457206} \\
\bottomrule
\end{tabular}
\end{table}

\subsection{Qualitative Analysis}
Figure~\ref{fig:qual_progression} visualizes the progressive masks produced by different stages of the pipeline. Across the six DOST scenes, RAFT provides the initial motion response, but it is easily affected by turbulence-induced pseudo-motion and often activates background regions. DINOv2 semantic priors help suppress part of these noisy responses by favoring semantically coherent object regions, while skip-frame motion estimation makes weak or slow displacement more visible. ViBe background modeling provides an additional anomaly cue, which is useful when moving targets appear as persistent deviations from the background. After multi-signal fusion, SAM2 box-prompt refinement converts coarse and fragmented proposals into more object-level masks with cleaner boundaries.

The experiment results also show the remaining failure cases. Extremely small or briefly visible targets, such as the bird sequence, are still difficult because the object can be confused with high-contrast background structures or local turbulence artifacts. Since the whole pipeline is training-free and relies on fixed manually selected parameters, future improvements may come from adaptive thresholding, object-level temporal tracking, and lightweight task-specific adaptation.

\FloatBarrier

\section{Conclusion}
We present a training-free multi-signal dynamic object segmentation system for the CVPR 2026 UG2+ DOST challenge. The framework combines pretrained RAFT optical flow, DINOv2 semantic priors, skip-frame motion estimation, ViBe background modeling, manually calibrated proposal fusion, and pretrained SAM2 box-prompt refinement. By using these cues jointly, the system reduces its dependence on any single motion response and produces temporally more stable masks for long-range videos affected by atmospheric turbulence. In the official leaderboard evaluation, the final system achieves 0.425041 mIoU and 0.457206 mDice on the labeled DOST validation subset. Since no DOST-specific training is performed, there remains room for improvement through adaptive proposal generation and task-specific fine-tuning of foundation segmentation models.

\end{document}